\documentclass{article}
\usepackage{graphicx}
\usepackage[numbers]{natbib}
\usepackage{booktabs} 
\usepackage{array} 
\usepackage{caption} 
\usepackage{tabularx}
\usepackage{amssymb}
\usepackage{multirow}

\usepackage{xcolor}
\usepackage{pifont}
\usepackage{makecell}
\usepackage{seqsplit}
\newcommand{\model}[1]{\texttt{\seqsplit{#1}}}

\newcommand{\cmark}{{\ding{51}}}
\newcommand{\xmark}{\textcolor{gray}{\ding{55}}}

\usepackage{amsmath}

\title{LLM Jaggedness Unlocks Scientific Creativity}
\date{May 2026}

\author{
Shray Mathur, J. Anibal Boscoboinik, \\
Esther H. R. Tsai, and Kevin G. Yager \\
\small Center for Functional Nanomaterials, Brookhaven National Laboratory \\
\small Upton, NY 11973, USA
}

\begin{document}

\maketitle

\section{Abstract}

As artificial intelligence advances, models are not improving uniformly. Instead, progress unfolds in a jagged fashion, with capabilities growing unevenly across tasks, domains, and model scales. In this work, we examine this dynamic jaggedness through the lens of scientific idea generation. We introduce SciAidanBench, a benchmark of open-ended scientific questions designed to measure the scientific creativity of large language models (LLMs). Given a scientific question, models are asked to generate as many unique and coherent ideas as possible, with the total number of valid responses serving as a proxy for creative potential.
Evaluating 19 base models across 8 providers (30 total variants including reasoning versions), we find that jaggedness manifests both across models and within models.  First, in a cross-task comparison between general and scientific creativity, improvements in general creativity do not translate uniformly to scientific creativity, revealing divergent capability profiles across models. Second, at the prompt level, stronger models do not improve uniformly; instead, they exhibit high variability, with bursts of creativity on some questions and limited performance on others. Third, at the domain level, individual models display uneven strengths across scientific subfields, reflecting fragmented internal capability profiles. Finally, we show that this jaggedness can be harnessed. We explore mechanisms of \textit{inference-time compute}, \textit{knowledge pooling}, and \textit{brainstorming} to combine models effectively and construct meta-model ensembles that outperform any single model. Our results position jaggedness not as a limitation, but as a resource, a structural feature of AI progress that, when understood and leveraged, can amplify LLM-driven scientific creativity.

\section{Introduction}

Recent research has revealed that progress in artificial intelligence is not uniform across capabilities but rather follows a \textit{jagged frontier} \cite{dell2023navigating}: an uneven landscape of competence. Although the notion of \textit{jaggedness} has only recently been introduced explicitly, multiple studies implicitly reveal this phenomenon across distinct dimensions of model characteristics. 
Diverse forms of jaggedness have been identified (Table~\ref{tab:jaggedness}), including: across tasks, where AI assistance improves human performance in some task types while worsening it in others \cite{dell2023navigating}; across model scale, where larger models may perform worse on certain benchmarks \cite{McKenzie2023-gf}; across reasoning effort, where allocating more computational steps can lead to declining accuracy \cite{Gema2025-bo}; and across problem complexity, where performance can collapse as task difficulty increases \cite{Shojaee2025-es}. More recently, behavioral jaggedness has been observed, where small differences in model alignment or value priors produce large behavioral divergences \cite{Zhang2025-bf}. Together, these results suggest that AI competence forms a highly irregular surface, one that varies across scale, compute, complexity, and specification. A systematic characterization of such jaggedness remains at an early stage.

\begin{table}[h]
\centering
\caption{Summary of different dimensions of jaggedness in AI capabilities across recent literature.}
\label{tab:jaggedness}
\renewcommand{\arraystretch}{1.3} 
\begin{tabularx}{\textwidth}{@{}>{\raggedright}p{2.5cm}>{\raggedright\arraybackslash}p{2.5cm}X@{}}
\toprule
\textbf{Paper} & \textbf{Jaggedness Dimension} & \textbf{Description} \\
\midrule
Dell'Acqua et al. \cite{dell2023navigating} & Task Frontier & AI performance varies sharply across task types---high gains within the frontier but degradation on tasks beyond it. \\
McKenzie et al. \cite{McKenzie2023-gf} & Model Scale & Increasing model size can reduce performance on certain tasks, revealing non-monotonic scaling effects. \\
Gema et al. \cite{Gema2025-bo} & Reasoning Depth & Allowing more reasoning steps or compute time can degrade accuracy, showing limits to deeper reasoning. \\
Shojaee et al. \cite{Shojaee2025-es} & Problem Complexity & Model accuracy collapses abruptly beyond certain complexity thresholds, exposing cliffs in reasoning ability. \\
Zhang et al. \cite{Zhang2025-bf} & Model Behaviour & Small differences in model specifications or normative values cause large behavioral divergences. \\
\bottomrule
\end{tabularx}
\end{table}

Building on these observations, we aim to study this jaggedness more explicitly and formally through the lens of scientific creativity. Creativity is an ideal measure for revealing differences because it is open-ended and generative, giving models the freedom to explore solution spaces. We build on AidanBench \cite{McLaughlin_undated-yd}, a benchmark designed to evaluate general creativity of LLMs via sustained idea generation to open-ended questions. We adapt the methodology to the scientific domain and present SciAidanBench \cite{mathur2025sciaidanbench}, a benchmark dataset of 155 open-ended scientific questions, spanning 6 scientific domains. 

We focus on scientific creativity for studying jaggedness for several reasons. Scientific creativity is closely related to general creativity but with the stricter constraints of scientific reasoning and coherence. Because both tasks share a similar formulation, where models are asked to generate as many valid and meaningful responses as possible, they provide a natural basis for comparison. This allows us to study jaggedness across tasks (general vs scientific creativity), revealing how model capabilities may shift under similar but different task types. The hierarchy of scientific sub-domains offers a way to evaluate jaggedness across topics. Finally, the growing use of large language models (LLMs) in scientific discovery \cite{Lu2024-jq, Yu2024-hq, Qi2024-hy}, from literature review \cite{shi2023towards, hsu2024chime} to experimental design and operation \cite{wu2023autogen, huang2024crispr, mathur2025vision}, makes this investigation timely.  Understanding jaggedness in this context not only sheds light on the unevenness of model capabilities but also offers insights that could be leveraged to smoothen out these jagged edges and enhance the scientific capabilities of AI systems.

Previous efforts to improve LLM diversity include either computationally intensive methods like training interventions \cite{chung2025modifying, zhou2023sotopia}, decoding strategies \cite{chang2025real, nguyen2024turning}, or lightweight prompting methods \cite{Zhang2025-bf}. Specifically there have been efforts to improve scientific idea generation capabilities of LLMs \cite{Su2024-ob, Radensky2024-ay, Baek2024-ym} which rely on multi-llm systems. While these approaches have been shown to improve idea generation capabilities, they remain constrained by structured workflows that limit open-ended generation. Moreover, most multi-LLM systems typically use replicas of the same model, inheriting shared biases and limiting diversity.

Through \textit{SciAidanBench}, we provide a benchmark for evaluating scientific idea generation capabilities of LLMs. The benchmark consists of a diverse set of open-ended scientific questions drawn from across the sciences (physics, chemistry, nanoscience, neuroscience, biology and environmental science). Given a question such as ``Propose an experiment to detect a new fundamental force of nature'' or ``Propose a new method for delivering CRISPR machinery into specific cell types,'' models are prompted to generate as many unique ideas as possible, while coherently answering the query. This measures their ability to explore conceptual spaces, and acts as a proxy measure for the need for creatvitiy in science more generally. We use this measure to shed light on jaggedness of LLMs. While prior work has attributed jaggedness to factors such as data composition, training methodology, and post-training alignment, little attention has been given to how it might be mitigated or harnessed. Studying creativity in this setting provides a means to do so, enabling the identification and combination of strengths across models to smoothen out the jaggedness of this developing intelligence frontier.

\section{Methodology}

\subsection{Dataset}

We construct a dataset comprising \textbf{155 open-ended scientific questions} designed to evaluate the creative reasoning and sustained ideation capabilities of large language models (LLMs). Each question requires domain-relevant scientific reasoning and open-ended concept generation rather than simple retrieval of factual knowledge. Questions are phrased using templates such as ``\textit{Propose a method for\dots}'' or ``\textit{Suggest a way to\dots}'' or ``\textit{Design an experiment that could\dots}.'' This encourages models to produce creative ideas rather than descriptive or encyclopedic answers.

The questions were curated by researchers at the Center for Functional Nanomaterials (CFN) at Brookhaven National Laboratory, ensuring domain authenticity and scientific depth. The dataset spans \textbf{six broad scientific domains} --- \textit{Physics, Chemistry, Biology, Neuroscience, Nanoscience}, and \textit{Environmental Science}. Physics is further subdivided into \textit{Fundamental Physics, Astrophysics, Synchrotron Physics}, and \textit{Condensed Matter Physics}. The complete list of questions is provided in the Appendix.

\subsection{Models}

We evaluate 30 state-of-the-art large language model variants across 19 base models and 8 providers; the full model list is provided in Appendix.

\begin{itemize}
    \item \textbf{Local models:} (e.g., \texttt{LLaMA 3.3}, \texttt{Mistral}, \texttt{Phi 3.5}) are hosted locally via \texttt{Ollama} and executed on \texttt{NVIDIA H100 GPUs}.

\item \textbf{Cloud models:} (e.g., \texttt{GPT-4o}, \texttt{Claude-3.7-Sonnet}, \model{Gemini-2.5-Pro}) are accessed via their respective vendor APIs --- Anthropic via AWS Bedrock, OpenAI via Azure, and Google/DeepSeek via a commercial model router (\texttt{Abacus.ai}).
\end{itemize}

\subsection{Task Design}

The task design directly follows that of AidanBench \cite{McLaughlin_undated-yd}. For each question \( q \), the LLM is prompted to generate an initial idea \( r_1 \). The model is then iteratively re-prompted to produce additional ideas \( r_2, r_3, \dots \) (where it can see ideas it has generated so far). Each new idea must both coherently answer the question, and be novel relative to all earlier ideas.

To enforce these constraints, every generated response is evaluated using two components:

\begin{enumerate}
    \item \textbf{Embedding-based novelty (computed by an embedding model):}  
    We use a pretrained embedding model (\texttt{text-embedding-3-large}) to compute the cosine distance between the new response \( r \) and the previous responses \( r' \) and calculate a novelty score as shown in Table \ref{tab:metrics}

    \item \textbf{LLM-based checks (computed by the judge LLM \( J \), \texttt{o1-mini}):}  
    The judge LLM performs two evaluations as shown in Table \ref{tab:metrics}:
    \begin{itemize}
        \item \textit{LLM-based novelty:} The judge compares \( r \) with earlier responses and gives a similarity score for the new response with each previous response. 
        \item \textit{Coherence:} The judge rates how well the response answers the question and whether it is logically consistent (0–100 scale).
    \end{itemize}
\end{enumerate}

This separation ensures that novelty is evaluated through both semantic distance (via embedding check) and conceptual distinctiveness (via LLM check). The iterative generation process continues until the model produces an answer that falls below a novelty threshold (\( < 0.15 \)), or below a coherence threshold (\( < 15 \)). The benchmark task is designed such that there is no upper limit to the possible score; this enables investigation of jaggedness without artificial metric saturation.

\begin{table}[h]
\centering
\caption{Evaluation metrics used to assess novelty and coherence of model-generated responses. Here, $q$ denotes the input question, $r$ a candidate response, and $R = \{r_1, \dots, r_{t-1}\}$ the set of previously generated responses for the same question. The judge model $J$ is a large language model used to compute similarity scores between responses and to evaluate coherence on a fixed scale.}
\begin{tabular}{|l|p{8cm}|}

\hline
\textbf{Metric} & \textbf{Definition} \\
\hline
Embedding-based novelty &
$E(r, R) = 1 - \max_{r' \in R} \dfrac{e(r) \cdot e(r')}{\|e(r)\|\|e(r')\|}$, where $e(r)$ is the embedding of response $r$. \\
\hline
LLM-based novelty &
$L(r, R) = 1 - \max_{r' \in R} J_{\text{sim}}(r, r')$, where $J_{\text{sim}}$ is the similarity score from the judge model $J$. \\
\hline
Coherence &
$C(r) = J(q, r)$, evaluated by $J$ on a $[0,100]$ scale. \\
\hline
\end{tabular}

\label{tab:metrics}
\end{table}







\section{Results}

\subsection{Exploring Jaggedness}

\paragraph{Cross-task comparison:}

\begin{figure}[h]
  \centering
  \includegraphics[width=1\textwidth]{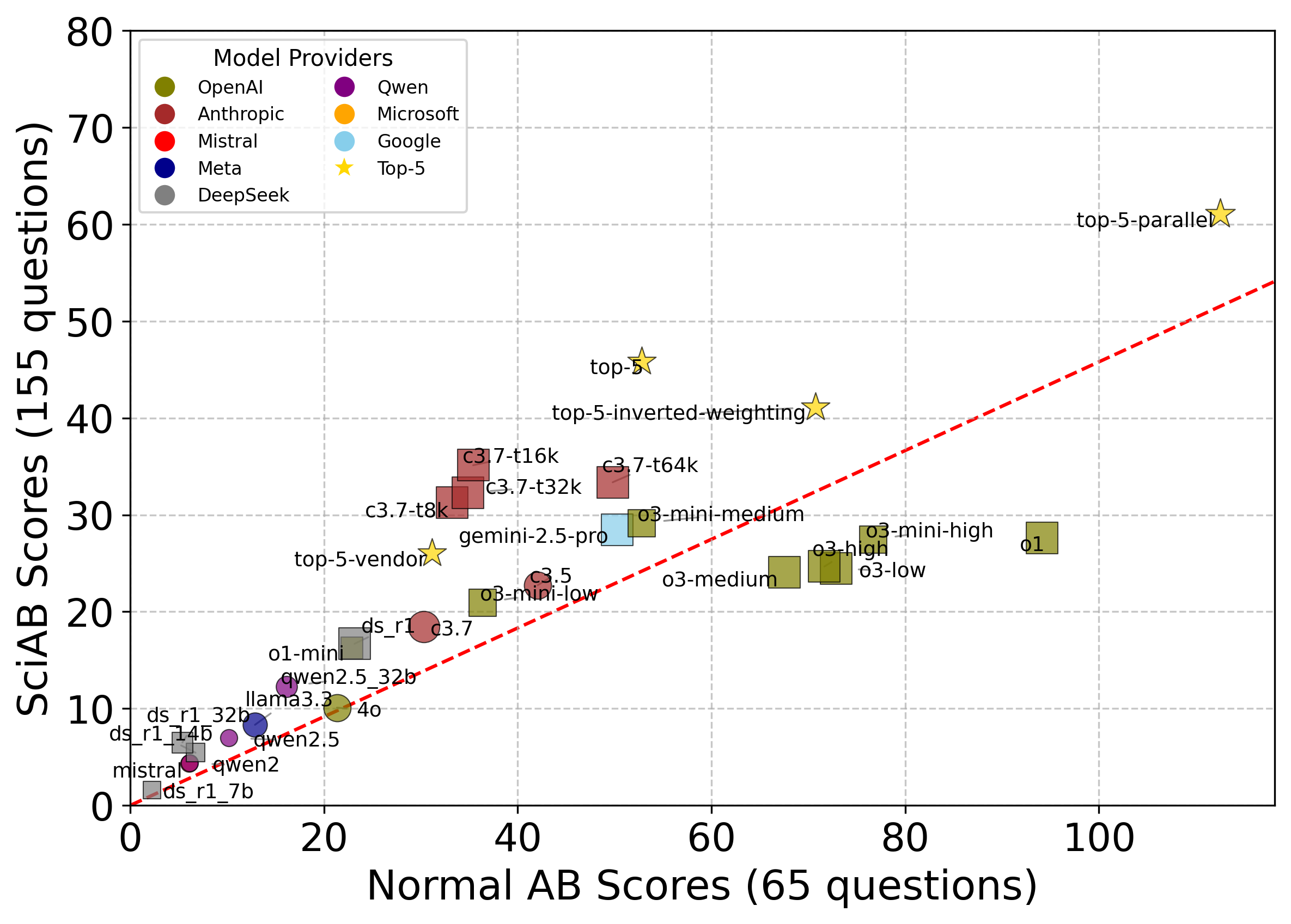}
  \caption{Average responses per question on general creativity (AidanBench, x-axis) versus scientific creativity (SciAidanBench, y-axis). The dashed red
  line shows the least-squares fit (computed over
  individual models only, excluding Top-5 ensemble configurations) through the origin ($y = 0.46x$). Circles denote non-reasoning models, squares
  denote reasoning models, and stars denote meta-model ensemble configurations defined in this work. Models are shown using
  short names; see Figure~\ref{fig:ribbon_plot} for the full names. The spread of models away from the trendline is one desmontration of jaggedness.}
  \label{fig:overall}
\end{figure}

When comparing model performance between general and scientific creativity, we find that improvements are not uniform across tasks. As shown in Figure~\ref{fig:overall}, performance on general (AidanBench) and science (SciAidanBench) tasks is positively correlated ($r = 0.76$). While this correlation is strong, notable deviations from the trend line reveal heterogeneity in model behavior. The slope of the best fit line ($y = 0.46x$) can be interpreted either as science generation being inherently more difficult, or our benchmark being more challenging. Notably, deviations of models from the average trendline highlights their differences. The ordering of models differs across the two tasks, revealing a form of cross-task jaggedness: models that rank highly in general creativity do not necessarily rank highly in scientific creativity, and vice versa. Capability gains occur along different axes for different models rather than scaling uniformly across creative domains.

Provider-level patterns illustrate additional asymmetry. OpenAI models are clustered and show strong improvements in general creativity but appear to plateau on scientific creativity, whereas Anthropic models outperform on scientific creativity relative to general creativity. These distinct regions occupied by different model families reflect characteristic capability profiles across tasks. These differences likely result from underlying differences in their construction, such as different training datasets or model architecture designs.

\begin{figure}[h]
  \centering
  \includegraphics[width=1\textwidth]{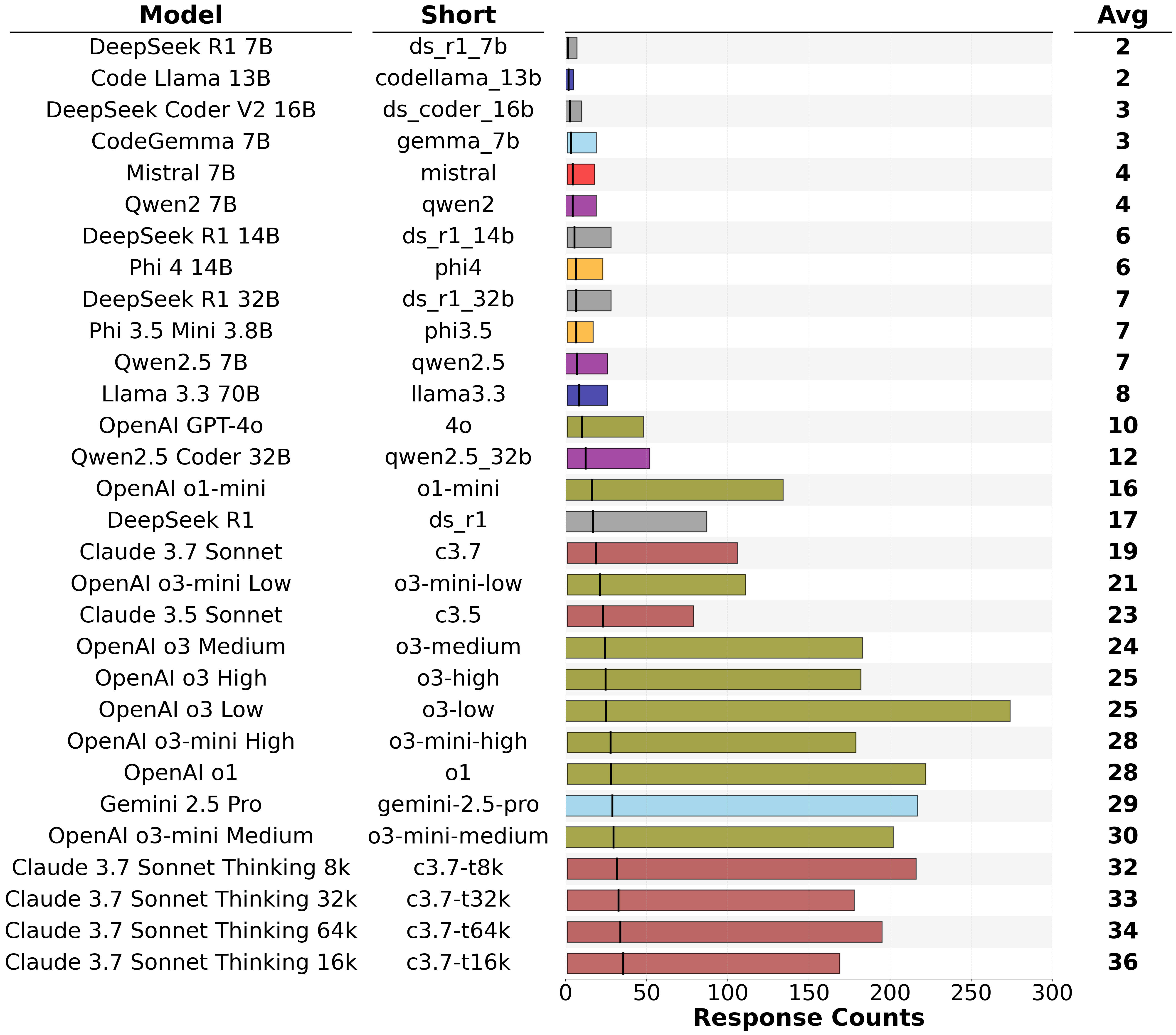}
  \caption{Response count ranges across models on SciAidanBench. Each row corresponds to a model, showing the full range of response counts (min–max) as a horizontal ribbon, with the mean indicated by a vertical black marker. Models are sorted by increasing average response count. The left columns list full and short model names, and the right column reports the mean value. It is clear that models exhibit significant variability across benchmark questions.}
  \label{fig:ribbon_plot}
\end{figure}

\paragraph{Prompt-level variability:}
Comparing models on SciAidanBench shows that gains in capability do not translate into uniform improvements across prompts. Instead, stronger models exhibit greater variability in the number of ideas they generate. This is evident in Figure~\ref{fig:ribbon_plot}. As model capability increases, the response-count ribbons widen substantially. High-performing models such as \model{o3-high}, \model{Gemini-2.5-Pro}, and \model{Claude-3.7-Sonnet-Thinking (8k-64k)} show ribbons that stretch from very low to very high response counts, indicating that even the strongest models struggle with certain prompts while producing bursts of creativity on others. In contrast, weaker models display narrow ribbons clustered near the low end, reflecting consistently limited output.

\begin{figure}[h]
  \centering
  \includegraphics[width=1\textwidth]{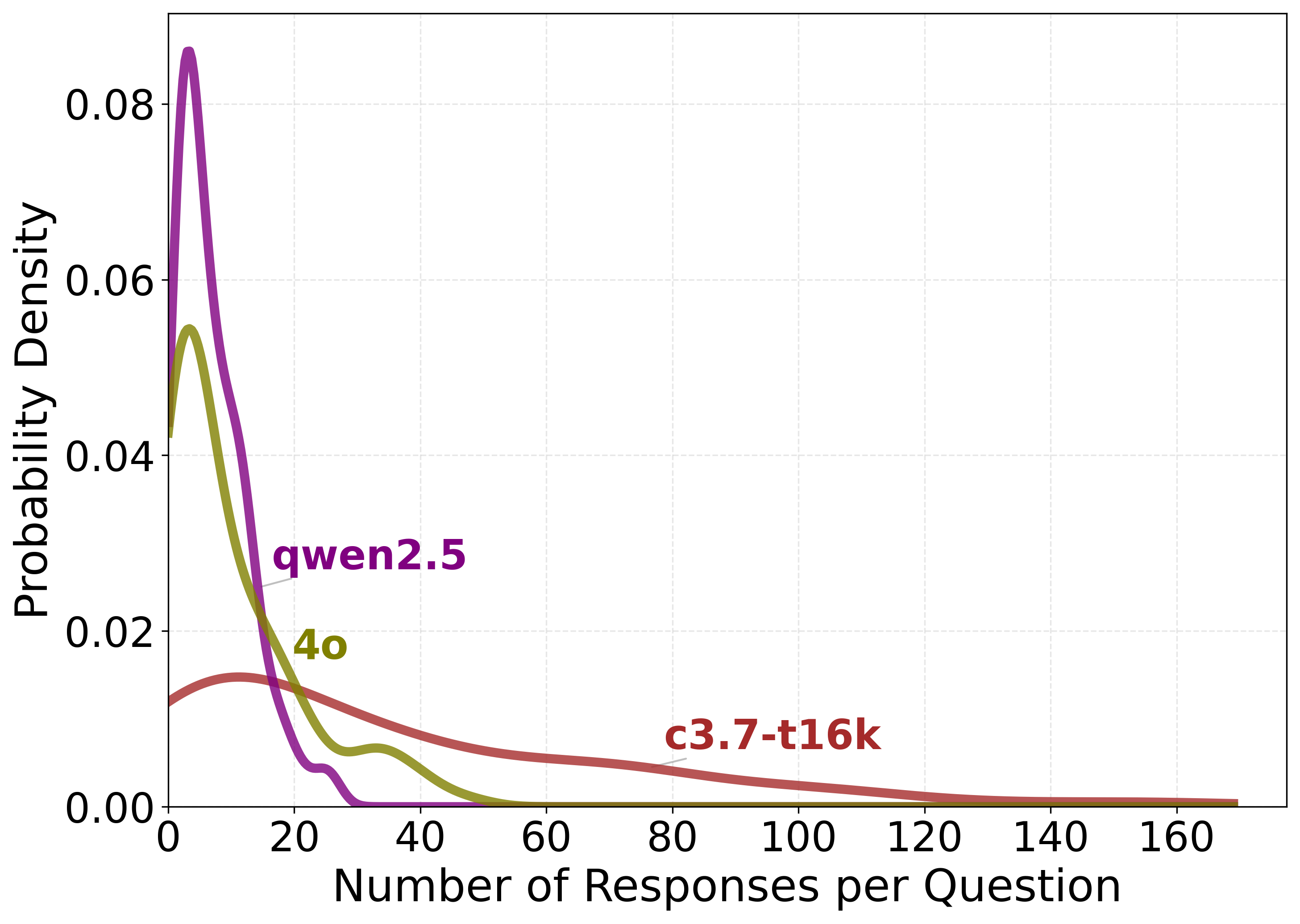}
  \caption{Probability density of response counts for three models of increasing capability (\protect\model{Qwen-2.5}, \protect\model{GPT-4o}, \protect\model{Claude-3.7-Sonnet-Thinking-16k}), showing the shift toward broader and heavier-tailed distributions as model strength increases. The higher average score of stronger models comes in large part from this tail of unusually high response count.}
  \label{fig:continous_dist}
\end{figure}

Figure~\ref{fig:continous_dist} illustrates this pattern in detail for \model{Qwen 2.5} (weaker), \model{GPT-4o} (mid-capability), and \model{Claude-3.7-Sonnet-Thinking-16k} (strongest). \model{Qwen 2.5} shows a tight, sharply peaked distribution centered at low counts, indicating uniformly constrained output. \model{GPT-4o} shifts rightward and broadens, developing heavier tails that reflect occasional high-score task completions. \model{Claude-3.7-Sonnet-Thinking-16k} expands this trend further, with a long tail capturing prompts that trigger exceptionally large idea sets. Together, these patterns show that to a large extent the improvement in mean score arises from the subset of highly successful responses; that is, high-capability models, on select questions, exhibit ``bursts'' of creativity. This again highlights the heterogeneous and unpredictable response behavior.

\begin{figure}[h]
  \centering
  \includegraphics[width=1\textwidth]{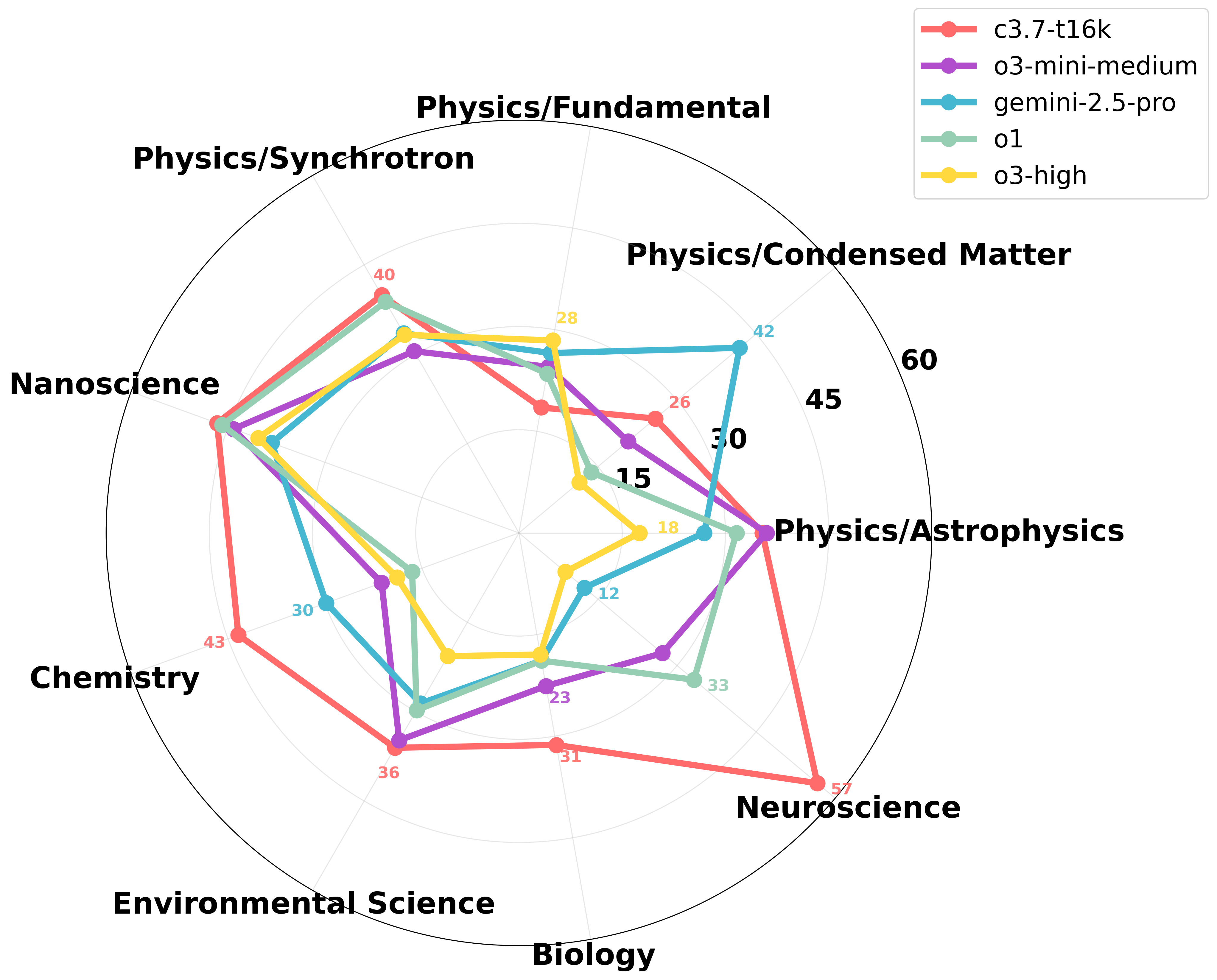}
  \caption{Domain-level response profiles for the top five distinct models on SciAidanBench, showing uneven strengths and weaknesses across scientific categories. Models are shown using short names; see Figure~\ref{fig:ribbon_plot} for the full names.}

  \label{fig:spider_plot_top5}
\end{figure}

\paragraph{Domain-level differences:}
Finally , we examine how jaggedness manifests within individual models across the scientific subdomains. Figure~\ref{fig:spider_plot_top5} shows the top five distinct models on SciAidanBench (selecting only the best-performing configuration where multiple reasoning variants of the same model exist) plotted across different scientific subdomains. Rather than displaying smooth or uniform profiles, these models exhibit distinct spikes, indicating uneven creative strengths across domains. For example, \model{Gemini-2.5-Pro} performs strongly on Condensed Matter Physics, but is weaker on Nanoscience, while \model{Claude-3.7-Sonnet-Thinking-16k} is strongest on Neuroscience but weaker on Fundamental Physics. This spikiness illustrates that creative ability is fragmented and domain-dependent. The frontier of scientific creativity thus remains irregular not only across models, but also within the internal capability profiles of models.

\paragraph{}

While jaggedness may appear as a limitation, it also presents an opportunity. If models differ in where their creative strengths lie, these differences can be exploited to build stronger composite systems. This motivates our exploration of meta-model ensembles.

\subsection{Exploiting Jaggedness: Meta-Model Ensembles}

The jaggedness observed in our analyses suggests that contemporary LLMs are not merely scaled versions of one another but are, in meaningful ways, \emph{different}. Prior work attributes such variation to factors including model scale, training data composition, architectural choices, and alignment strategies \cite{wei2022emergent, gao2020pile, fedus2022switch, ouyang2022training}; however, regardless of its origin, this divergence manifests as differences in what models know, how they reason, and the kinds of creative moves they tend to make. 


For creativity, particularly scientific creativity, this diversity is potentially advantageous. For human creativity, teams comprising different perspectives, knowledge bases, and thought processes often generate ideas that no individual would have produced alone \cite{hong2004groups}. If models likewise vary in how they approach creative problems, then their jaggedness represents an opportunity rather than a limitation. This raises a natural question: \emph{can we harness the diversity of LLMs to produce stronger creative performance than any single model can achieve on its own?} 

To explore this possibility, we designed ensemble meta-models that encourage models to think, build on one another’s outputs, and share their different knowledge bases. We explore three mechanisms through which ensembles can leverage  diversity to improve scientific creativity: \textit{inference-time compute}, \textit{knowledge pooling}, and \textit{brainstorming}. Inference-time compute increases creative exploration by simply allowing models to think more and explore more of the solution space. Knowledge pooling aggregates complementary domain expertise across systems, while brainstorming aims to enable iterative refinement as models build upon one another’s ideas. Each mechanism provides a distinct pathway for converting uneven capability profiles into collective creative strength.

To test these mechanisms, we construct ensemble configurations that implement these principles independently and in combination. Table \ref{tab:mechanisms} shows which mechanisms are used for different models. We then evaluate their performance on SciAidanBench to determine whether ensembles can outperform individual models and whether their improvements correspond to a smoothing of jaggedness in creative performance. We find that ensemble models score significantly higher on our benchmark, as compared to their constituent models (Figure~\ref{fig:overall}). This reframes jaggedness as a structural asset: by coordinating models that differ in their creative tendencies, meta-model ensembles can transform fragmentation into synergy, amplifying scientific idea generation beyond the limits of any single model.

\begin{table}[t]
\centering
\begin{tabular}{lccc}
\toprule
\textbf{Model} & \makecell{\textbf{Inference-Time} \\ \textbf{Compute}} & \textbf{Pooling} & \textbf{Brainstorming} \\
\midrule
\multicolumn{4}{l}{\textit{Individual Models}} \\
Claude-3.5-Sonnet            & \xmark & \xmark & \xmark \\
Claude-3.7-Sonnet-Thinking   & \cmark & \xmark & \xmark \\
\midrule
\multicolumn{4}{l}{\textit{Meta-Models}} \\
Router                       & \xmark & \cmark & \xmark \\
Top-5                        & \xmark & \cmark & \cmark \\
Top-5-Parallel               & \cmark & \cmark & \cmark \\
\bottomrule
\end{tabular}
\caption{Ensemble configurations and the mechanisms they implement.}
\label{tab:mechanisms}
\end{table}

\subsubsection{Inference-Time Compute}

A natural hypothesis for scientific creativity is that deeper or longer reasoning should expand the space of ideas a model can explore. Modern LLMs operationalize thinking via reasoning tokens, i.e, tokens generated in the intermediate chain-of-thought before producing final answers.  We investigate whether enabling this reasoning capability improves performance on SciAidanBench.

\begin{figure}[h]
  \centering
  \includegraphics[width=1\textwidth]{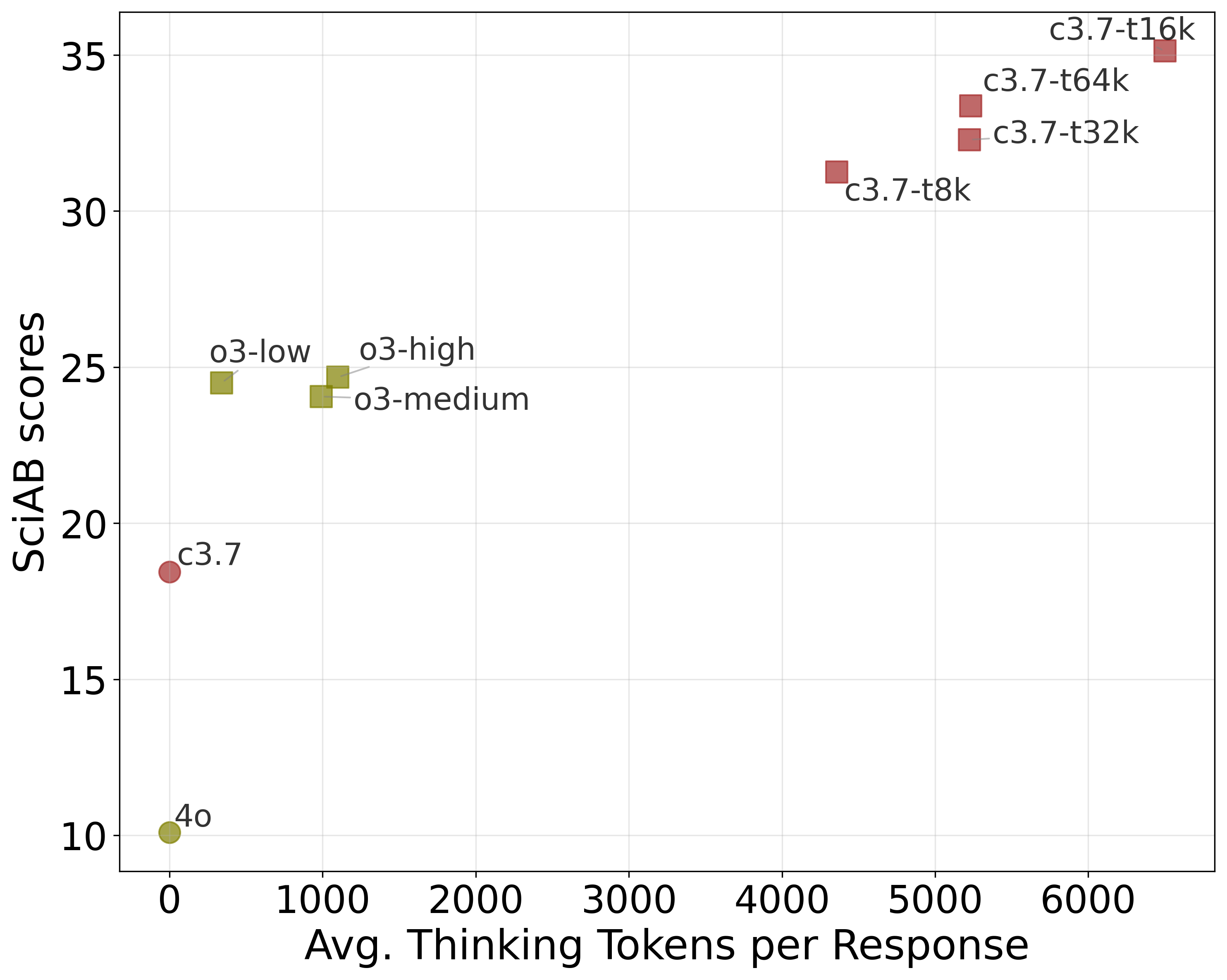}
  \caption{Relationship between average reasoning token usage and SciAidanBench scores. Olive markers represent OpenAI models and brown markers represent Anthropic models. Circles denote non-reasoning models, while squares denote reasoning models. Models are shown using short names; see Figure~\ref{fig:ribbon_plot} for the full names. In general, more inference-time compute leads to higher creativity scores.}

  \label{fig:thinking_tokens}
\end{figure}

Figure~\ref{fig:thinking_tokens} plots selected Anthropic and OpenAI models against the average number of reasoning tokens used per response. The Claude-3.7-Sonnet model family provides the most direct comparison as it can be run with or
without extended thinking, holding the base model constant. Without reasoning tokens, \model{Claude-3.7-Sonnet} scores around 18. Enabling extended thinking produces a substantial jump: the reasoning-enhanced variants score between 31 and 35, with
\model{Claude-3.7-Sonnet-Thinking-16k} achieving the highest performance. Notably, \model{Claude-3.7-Sonnet-Thinking-16k} also consumes the most reasoning tokens in practice ($\sim$6000 on average), despite having a lower allocated budget than the 32k and 64k variants, which use $\sim$5,000 tokens on average and score slightly lower. This suggests that a larger allocated budget does not guarantee deeper reasoning or better performance; what matters is the actual number of reasoning tokens consumed.

OpenAI's o3 model series does not have a non-reasoning variant, so we compare against \model{GPT-4o}, the strongest non-reasoning OpenAI model on SciAidanBench. The pattern is consistent: \model{GPT-4o} scores around 10, while o3 variants score more than double that ($\sim 25$). Within the o3 family, the low, medium, and high reasoning-effort variants cluster tightly in both token usage and score, indicating a plateau in returns from additional reasoning effort.

These results support inference-time compute as a mechanism for improving scientific creativity: across both model families, the transition from non-reasoning to reasoning models yields consistent and substantial gains. Among reasoning variants,
performance tracks actual token consumption rather than allocated budget, and returns diminish beyond a certain depth of reasoning. It is increasingly understood that inference-time compute yields improvements in response quality, especially for tasks with verifiable rewards, as models have the opportunity to test ideas, backtrack, and so on. Our results show that similar arguments apply to constrained idea generation, as models are able to iteratively propose ideas and weed out bad ones (incoherent or repetitive) before committing to a reply. We expect LLM creativity to generally increase, as future models leverage even more inference-time compute.

\subsubsection{Knowledge Pooling}

Knowledge pooling is motivated by the observation that different LLMs possess different knowledge bases, reasoning styles, and domain specializations. Jaggedness across scientific categories shows that no single model is uniformly strongest. If these strengths reflect genuine differences in underlying knowledge or familiarity with particular subfields, then allowing models to ``pool'' their knowledge, by letting each one contribute, should improve overall scientific creativity. 

The \textit{Router} ensemble provides a simple demonstration of this idea. For each domain in SciAidanBench, we identify the model with the highest score in that category and route all questions from that domain to that model. This creates a meta-model composed of domain specialists, each handling the questions for which it is best suited.

As shown in Figure~\ref{fig:spider_plot}, the Router surpasses or matches the strongest individual model (\model{Claude-3.7-Sonnet-Thinking-16k}) in every domain, illustrating that pooling distributed domain knowledge directly converts jaggedness into improved scientific creativity. By allowing each model to contribute where its knowledge is deepest, knowledge pooling can provide a straightforward and effective mechanism for leveraging diversity across LLMs.


\subsubsection{Brainstorming}

The brainstorming mechanism is designed to enable cross-model idea refinement by allowing different models to contribute sequentially to a growing chain of responses. The strategy is simple: we begin by selecting a set of $k$ unique models, and then, for each answer generation step, we sample one model from this set to produce the next idea. Because each model sees all previously generated ideas, including those written by other models, this approach allows them to build upon one another’s contributions. In effect, the mechanism lets heterogeneous reasoning styles interact (indirectly) during scientific ideation. This indirect context-sharing mechanism differs from more explicit multi-agent debate frameworks, in which models directly critique and rebut one another across multiple rounds \cite{du2024improving, liang2024encouraging}. We adopt the former as a minimal and efficient means of testing the brainstorming hypothesis.

We use $k=5$ models and explore two ways of choosing this set. The first configuration, which we call \textit{Top-5 Models}, selects the five highest-scoring unique models on SciAidanBench. For this set, we study two sampling schemes: one in which models are sampled in proportion to their SciAidanBench performance (better models contribute more), and a second ``inverted'' scheme in which lower-scoring models within the top five are sampled more frequently. As shown in Figure~\ref{fig:overall}, the performance-proportional version produces a clear improvement over all individual models indicating that cross-model idea building among strong contributors yields genuinely higher scientific creativity. The inverted-weighting variant performs slightly worse but also is better than any individual model on SciAidanBench.

The second configuration, \textit{Top-5 Vendors}, selects the top 5 performing model but each from a different provider, to enforce any architectural and training diversity. For this ensemble, we use only the performance-proportional sampling scheme. As shown in Figure~\ref{fig:overall}, this configuration falls below the strongest individual model, appearing near mid-performing models rather than forming a superior ensemble. This indicates that increasing vendor diversity alone is insufficient; effective brainstorming requires not just heterogeneous models, but contributors that meet a certain capability level whose reasoning styles can interact productively.

\subsubsection{Combining Mechanisms}

The \textit{Top-5-Parallel} ensemble integrates all three mechanisms—inference-time compute, knowledge pooling, and brainstorming—by coordinating the top five models in parallel during each step of idea generation. The procedure operates as follows:

\begin{enumerate}
    \item 
    For each question, all five models are prompted simultaneously, each producing a candidate idea for the current step.

    \item
    Before generating their candidate for a given step, each model is shown the chain of previously \emph{selected} ideas.  
    These ideas may have come from any model in earlier steps and are anonymized, so models only see the content, not its source.

    \item
  The five candidate responses generated are randomly shuffled and presented to an independent \emph{idea selector}: \model{Claude-3.7-Sonnet}. The idea selector receives the original question, the full chain of
  previously accepted ideas, and the shuffled set of candidates. It then selects the single candidate it deems the most promising continuation of the chain. Crucially, the idea selector does not apply the SciAidanBench scoring rubric; its role is purely
   to identify the best next candidate idea.

    \item 
    The selected idea is appended to the chain and then evaluated by the scoring pipeline (embedding-based novelty, LLM-based novelty, and coherence), exactly as in the single-model setting.
\end{enumerate}

\begin{figure}[h]
  \centering
  \includegraphics[width=1\textwidth]{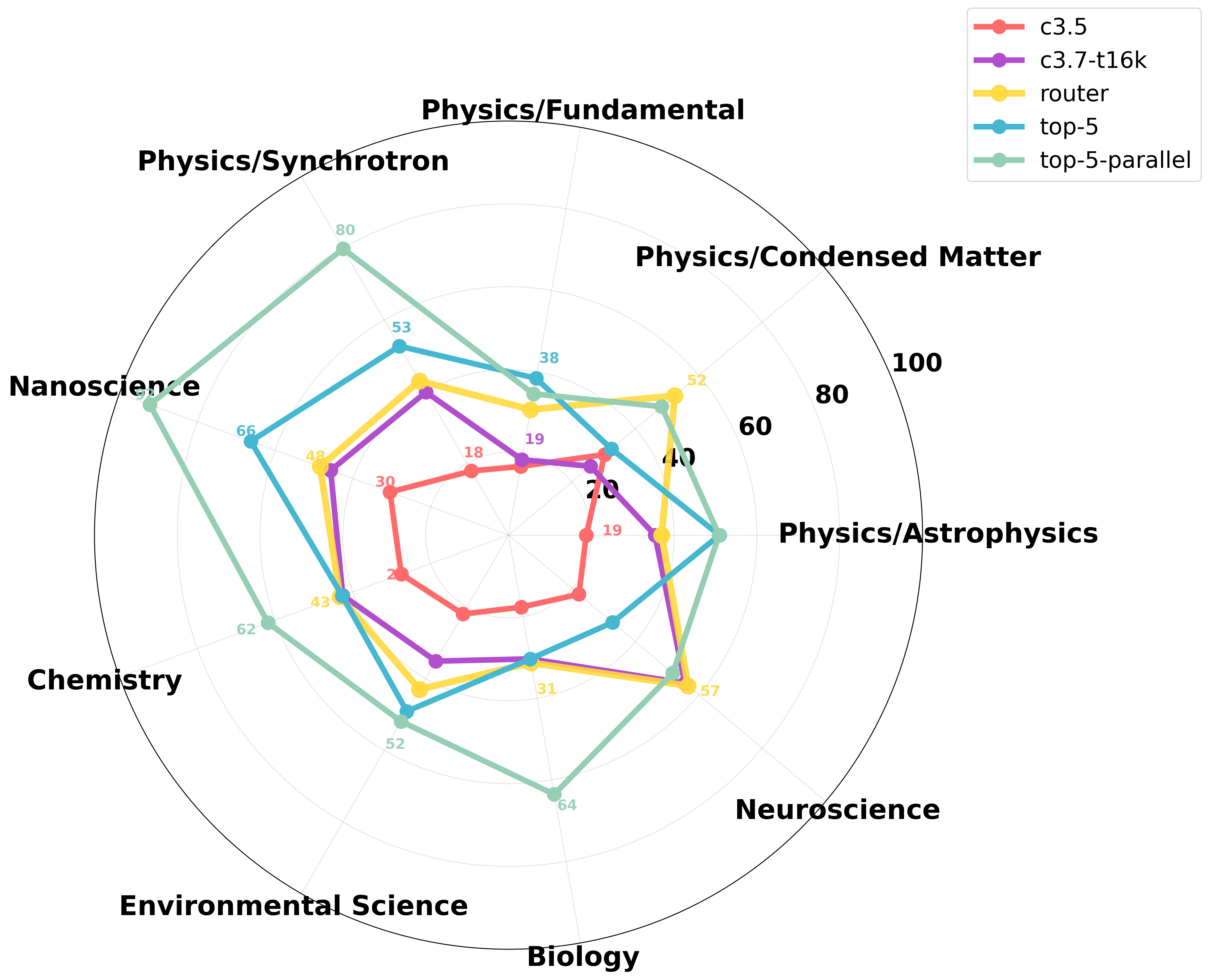}
 \caption{Domain-level performance across SciAidanBench subfields for the strongest individual models (\protect\model{Claude-3.5-Sonnet}, \protect\model{Claude-3.7-Sonnet-Thinking}) and for the three ensemble strategies (Router, Top-5, Top-5-Parallel). Meta-models, by combining individual models synergistically, outperform their constituent models.}

  \label{fig:spider_plot}
\end{figure}

Top-5-Parallel outperforms all individual models and ensemble variants on both general and scientific creativity (Figure ~\ref{fig:overall}) , and also achieves the highest scores in nearly every scientific domain (Figure ~\ref{fig:spider_plot}). It demonstrates that coordinating multiple mechanisms yields gains that exceed those achievable by any single model or single mechanism alone.

The success of these rather straightforward meta-model methods suggests a general approach for improving LLM performance, as long as there remains some heterogeneity in their design and capabilities. Specifically, models can be combined, allowing them to interact and discuss, and thereby yielding higher quality responses. While we have tested for scientific creativity here, we expect similar gains to observed across a broad range of tasks.

\section{Conclusion}

Using SciAidanBench we demonstrate that creative capabilities vary significantly across models, across tasks, across prompts, and across scientific subfields within a single model. Improvements in general creativity do not necessarily translate to scientific creativity, stronger models exhibit greater variance rather than uniform gains, and individual models display fragmented domain-specific strengths. Together, these results reveal scientific idea generation as an irregular and jagged capability surface. We view this as a specific manifestation of general model jaggedness.

Beyond characterizing this unevenness, we show that this jaggedness is not merely a limitation to be smoothed away, but a resource that can be exploited to improve creativity capabilities. By leveraging inference-time compute, pooling domain strengths, and enabling cross-model brainstorming, meta-model ensembles outperform their strongest constituent models. The success of meta-model ensembles, and the mechanisms that underpin them, demonstrate that expanding the creative surface a model can explore is a direction worthy of further investigation. A promising direction is to deliberately perturb the prompt context to encourage access to more of a model's jagged creative surface. Injecting random auxiliary content, unrelated facts, or alternative perspectives may nudge models into qualitatively different generations and expose creative modes that are otherwise underused. A possible instantiation of this idea within SciAidanBench would be to seed a model's context with concepts or findings drawn from a different scientific domain than the question being posed. For instance, providing a random idea from condensed matter physics as auxiliary context when the model is answering a neuroscience question. The hypothesis is that cross-domain seeding may prompt the model to draw unexpected analogies, surface latent connections, and explore regions of its answer space that standard prompting leaves untouched \cite{wu2024generative}. This connects to recent work showing that infusing random concepts into prompts helps increases output diversity \cite{agrawal2026addressing}, and that multilingual prompting can unlock generation diversity \cite{wang2025multilingual}. Another direction is scaling collaborative ensembles. One question is team size: for a given setup, how many models should be used to get the best performance? A second question is team structure: we only study horizontal teams, where all models are at the same level in the hierarchy. Other structures, including hierarchical or mixed setups, may allow larger teams to work together more effectively.


SciAidanBench serves a dual role: as a benchmark for scientific ideation and as a framework for revealing and harnessing structural heterogeneity in model capabilities. More broadly, our findings suggest a shift in how progress in AI should be interpreted. Rather than expecting monotonic improvements from scaling or single-model optimization, future gains, particularly in open-ended scientific idea generation, may increasingly arise from systems that explicitly recognize, measure, and orchestrate jagged intelligence.

\section*{Acknowledgments}

We thank Noah van der Vleuten for helpful discussions and copyediting assistance. This research was performed by the Center for Functional Nanomaterials (CFN), which is a U.S. Department of Energy Office of Science User Facility, at Brookhaven National Laboratory under Contract No. DE-SC0012704.

\section*{Data and Code Availability}

All data and code used in this work are publicly available at: \\
\texttt{https://github.com/CFN-softbio/SciAidanBench}

\bibliographystyle{unsrtnat}
\bibliography{paperpile, references}

\end{document}